\documentclass[10pt,twocolumn,letterpaper]{article}

\usepackage[pagenumbers]{cvpr} 

\definecolor{cvprblue}{rgb}{0.21,0.49,0.74}
\usepackage[pagebackref,breaklinks,colorlinks,allcolors=cvprblue]{hyperref}

\def\paperID{16963} 
\def\confName{CVPR}
\def\confYear{2026}

\title{Radioactive 3D Gaussian Ray Tracing for Tomographic Reconstruction}

\author{Ling Chen\\
Independent Researcher\\
\and
Bao Yang\\
Southern Medical University\\
}

\begin{document}
\maketitle
\begin{abstract}
3D Gaussian Splatting (3DGS) has recently emerged in computer vision as a promising rendering technique. By adapting the principles of Elliptical Weighted Average (EWA) splatting to a modern differentiable pipeline, 3DGS enables real-time, high-quality novel view synthesis. Building upon this, R\textsuperscript{2}-Gaussian extended the 3DGS paradigm to tomographic reconstruction by rectifying integration bias, achieving state-of-the-art performance in computed tomography (CT). To enable differentiability, R\textsuperscript{2}-Gaussian adopts a local affine approximation: each 3D Gaussian is locally mapped to a 2D Gaussian on the detector and composed via alpha blending to form projections. However, the affine approximation can degrade reconstruction quantitative accuracy and complicate the incorporation of nonlinear geometric corrections. To address these limitations, we propose a tomographic reconstruction framework based on 3D Gaussian ray tracing. Our approach provides two key advantages over splatting-based models: (i) it computes the line integral through 3D Gaussian primitives analytically, avoiding the local affine collapse and thus yielding a more physically consistent forward projection model; and (ii) the ray-tracing formulation gives explicit control over ray origins and directions, which facilitates the precise application of nonlinear geometric corrections, e.g., arc-correction used in positron emission tomography (PET). These properties extend the applicability of Gaussian-based reconstruction to a wider range of realistic tomography systems while improving projection accuracy.
\end{abstract} 
\section{Introduction}
\label{sec:intro}

Tomographic imaging modalities, such as Computed Tomography (CT) \cite{hounsfield1980computed,kak2001principles,withers2021x} and Positron Emission Tomography (PET) \cite{bailey2005positron,bar2000pet,townsend2004physical}, are widely used for medical diagnosis, non-destructive industrial testing, and scientific research. The primary objective is to accurately reconstruct the internal three-dimensional anatomical structure or functional distribution of a subject. For CT, classical analytical algorithms such as Feldkamp–Davis–Kress (FDK) \cite{feldkamp1984practical} are computationally efficient but suffer severe quality degradation under sparse projections, high noise levels, or physical effects like beam hardening. For PET, statistical iterative methods, e.g., Ordered Subsets Expectation Maximization (OSEM) \cite{hudson1994accelerated,rapisarda2010image}, model the imaging physics and the Poisson measurement noise, yielding improved image quality as compared to analytic methods; however, their performance depends strongly on the accuracy of the system matrix, which can be difficult to obtain or expensive to store.

Recently, implicit neural representations from computer vision, such as Neural Radiance Fields (NeRF) \cite{mildenhall2021nerf,zha2022naf,cai2024structure,LonYux_PDINR_MICCAI2025}, introduced a new paradigm for 3D reconstruction by learning continuous mappings from spatial coordinates to radiance and density. However, these methods typically require long training and rendering times, limiting their practicality in clinical or industrial settings that demand fast turnaround. A major step toward practicality is 3D Gaussian Splatting (3DGS) \cite{kerbl20233d,laine2020modular,yu2024mip,talegaonkar2025volumetrically,thirgood2025hypergs}, which replaces implicit MLPs with many explicit, optimizable 3D Gaussian primitives. Each primitive is parameterized by position, orientation, scale, opacity and view-dependent coefficients; a differentiable rasterization pipeline projects these Gaussians to the image plane and blends them via alpha compositing by combining classical Elliptical Weighted Average (EWA) splatting \cite{zwicker2002ewa,zwicker2001surface} with gradient-based optimization. 3DGS achieves much faster training and real-time novel-view synthesis while retaining high visual quality.

Motivated by these capabilities, researchers have adapted Gaussian representations for tomographic reconstruction. R\textsuperscript{2}-Gaussian \cite{zha2024r} is a notable example that adapts 3DGS for CT reconstruction by addressing an integration bias in the original splatting-based projection. In that framework, the scene is represented by Gaussians whose attributes correspond to physical quantities, e.g., attenuation, and the Gaussian parameters are optimized so that rasterized projections match measured scan data. R\textsuperscript{2}-Gaussian demonstrates promising high-resolution reconstructions on CT by introducing a projection model more consistent with respect to views. Similar to 3DGS, R\textsuperscript{2}-Gaussian adopts a local affine approximation: a 3D Gaussian is locally collapsed into a 2D Gaussian on the detector via an affine mapping, and contributions are accumulated for standard alpha compositing. This approximation simplifies gradient computation and improves efficiency, but it also can degrade integration exactness—an issue for quantitative imaging that demands high accuracy. Moreover, some real imaging systems exhibit geometric characteristics, e.g, cylindrical arrangement of PET detectors \cite{turkington2001introduction}. In PET, coincident gamma detections define Lines of Response (LORs) that may require arc correction \cite{buchert2000performance} to reconcile the detector geometry with reconstruction assumptions. Methods relying on local affine splatting encounter difficulties in incorporating such non-uniform bin spacing in the sinogram precisely, which limits their applicability in systems where accurate geometric modeling is essential.

\begin{figure*}[t]
  \centering
   \includegraphics[width=0.8\linewidth]{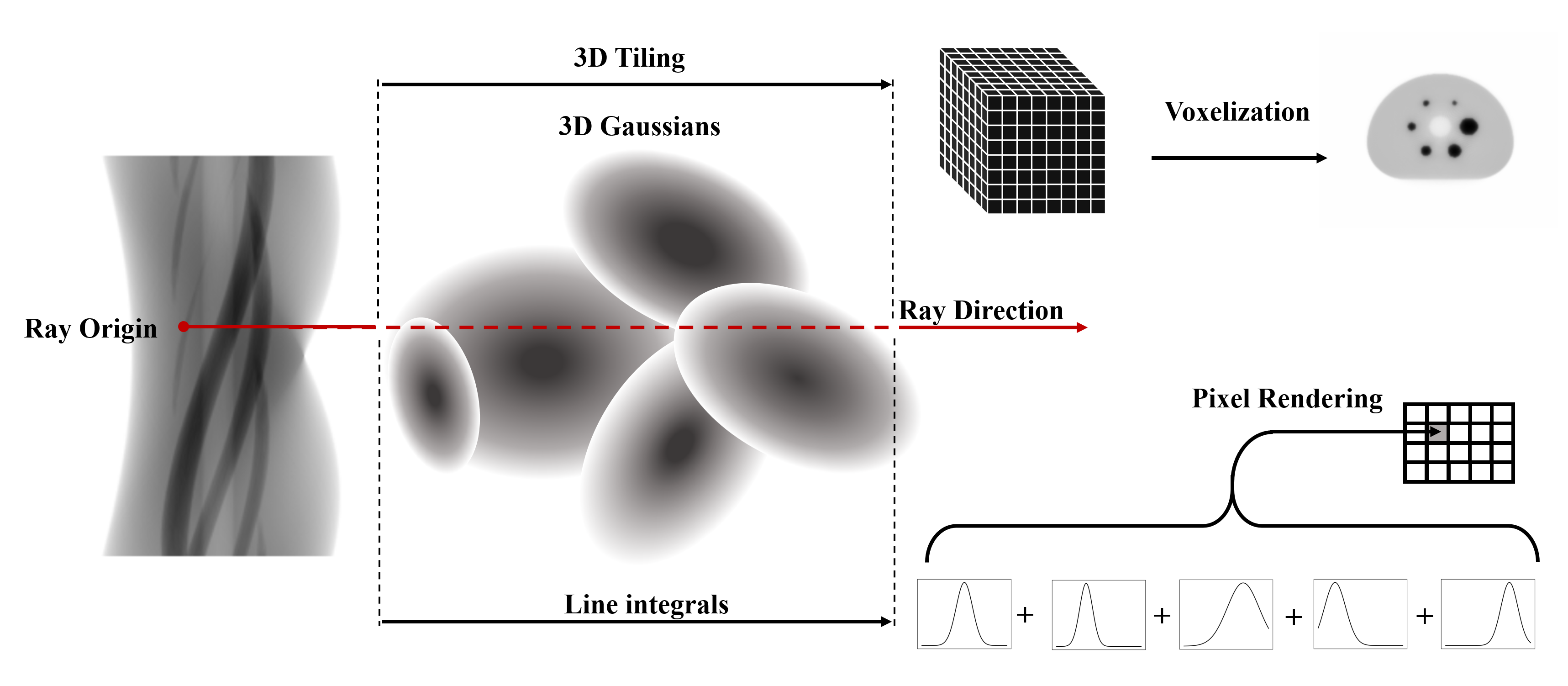}
   \caption{Overview of 3D Gaussian ray tracing for tomographic reconstruction. Given the scanner geometry, each ray’s origin and direction are defined. Along each ray, we compute the pixel value by analytically integrating the contributions of all 3D Gaussian primitives the ray encounters. By optimizing the Gaussian parameters so that the rendered projections match the measured projections, we obtain a tomographic reconstruction by voxelizing the optimized Gaussians.}
   \label{fig:intro}
\end{figure*}

To address these limitations, we propose a tomographic reconstruction framework based on 3D Gaussian ray tracing \cite{moenne20243d,condor2025don}. The framework is straightforward for tomographic reconstruction as shown in \cref{fig:intro}. Instead of projecting into 2D screen space via local affine approximation, our approach returns to the physical primitives of tomographic imaging—ray tracing and exact line integrals—bringing two main advantages:

\noindent\textbf{Accurate projective model.\quad} We derive and implement an analytical expression for the exact line integral of a ray through a 3D Gaussian primitive, eliminating the local affine approximation and improving numerical consistency.

\noindent\textbf{Compatibility with different scanner geometries.\quad} 3D Gaussian Ray tracing gives explicit control over ray origins and directions, allowing us to compatible with different scanner geometric characteristics precisely during ray setup and thus adapt the method to imaging systems such as PET.

The proposed framework is quantitatively and qualitatively evaluated on both PET and CT modalities. For PET, we perform (i) an analytical National Electrical Manufacturers Association (NEMA) phantom \cite{bao2009performance} simulation to evaluate reconstruction performance, (ii) a three-point-source Monte Carlo simulation to investigate the influence of arc versus non-arc correction, and (iii) qualitative assessment using realistic PET data. For CT, experiments are conducted on the same synthetic and real-world datasets as R\textsuperscript{2}-Gaussian, with reconstruction quality compared using 3D PSNR and 3D SSIM metrics.

\section{Related work}
\label{sec:RelatedWork}

\subsection{Tomographic Reconstruction}

The core objective of tomographic reconstruction algorithms is to reconstruct images of an internal structure from externally acquired projection data, primarily categorized into analytical and iterative methods. Analytical methods \cite{rit2009comparison}, represented by FDK, are based on rigorous mathematical models and are computationally efficient, serving as the standard algorithm for scenarios with complete, high signal-to-noise ratio data. Iterative methods \cite{vandenberghe2001iterative} employ statistical models, e.g., OSEM, approaching an optimal solution through successive iterations, offering superior advantages when handling incomplete or low signal-to-noise ratio data.

The choice between reconstruction algorithms for CT and PET is profoundly influenced by their underlying physical principles and data characteristics. CT measures the deterministic attenuation of X-rays, resulting in high-quality data, which is why FDK has long been its mainstream algorithm; iterative methods are primarily utilized in low-dose CT scenarios to suppress noise. In contrast, PET measures stochastic coincidence counts that follow a Poisson distribution, meaning the data is inherently characterized by low signal-to-noise ratio. This makes iterative methods the absolute dominant reconstruction technique for PET, as they effectively handle noise and allow for the incorporation of physical corrections.

\subsection{R\textsuperscript{2}-Gaussian}

R\textsuperscript{2}-Gaussian introduces the concept of explicit Gaussian primitives, previously used in neural rendering, into the task of tomographic reconstruction. It proposes a Gaussian representation and forward projection pipeline adapted for the physics of X-ray imaging. Unlike the typical 3DGS, R\textsuperscript{2}-Gaussian re-derives the projection relationship from 3D anisotropic Gaussians to the 2D detector plane by addressing integration bias. This constructs a radiative Gaussian kernel formulation that is more consistent with radiative transfer. Based on this formulation, Zha et al. \cite{zha2024r} implemented a 3D Gaussian splatting frame work for tomographic reconstruction.

Regarding scaling integration bias, the core idea of R\textsuperscript{2}-Gaussian is to recover the 3D density volume by 3D Gaussian splatting. This approach corrects the density inconsistency in 3DGS, ensuring the contribution of a single Gaussian to a pixel remains consistent across different projection angles. To maintain efficiency, R\textsuperscript{2}-Gaussian is typically coupled with sparse acceleration strategies, such as frustum culling, to limit the number of primitives requiring alpha compositing.

\subsection{3D Gaussian Ray Tracing}

3D Gaussian Ray Tracing treats the scene as a volumetric collection of 3D Gaussian primitives \cite{moenne20243d,zhou2025lidar,gao2024relightable}. It directly computes ray-primitive intersections and evaluates the corresponding integrals along pixel rays, rather than splatting the Gaussians into screen space and then performing approximate blending. By constructing a Bounding Volume Hierarchy (BVH) \cite{meister2021survey} and leveraging high-performance GPU ray-tracing hardware, the method casts rays for each pixel and processes depth-ordered intersections in batches, while the sorting process can be waived if it applies to tomographic reconstruction.

Compared to splatting-based pipelines, ray tracing offers three key advantages: First, it can naturally and accurately evaluate the line integral and transmittance along the ray, thereby avoiding the assumption of non-overlapping Gaussians. Second, not only 3D Gaussian kernel but also a more compact kernel, e.g., Epanechnikov kernel \cite{chu2017discrete}, can be applied to ray tracing \cite{condor2025don}. Third, we can define the ray origin and direction very flexible, which make this method compatible with different geometric characteristics of scanner. Thus, for application scenarios requiring more physically consistent projection or complex geometric correction, 3D Gaussian Ray Tracing demonstrates significant advantages over traditional 3D Gaussian Splatting in terms of numerical accuracy and flexibility. 
\section{Method}
\label{sec:Method}

In this section, we outline the preliminaries of R\textsuperscript{2}-Gaussian and the limitations we encounter in PET image reconstruction. After that, we present a detailed derivation of 3D Gaussian ray tracing for tomographic reconstruction.

\noindent\textbf{R\textsuperscript{2}-Gaussian.\quad} The core innovation in R\textsuperscript{2}-Gaussian lies in its derivation of a density-consistent integral for Gaussians, which reveals a previously unidentified integration bias in the standard 3DGS formulation. This bias emerges from approximations in the Gaussian projection integral, leading to significant inconsistency in density retrieval. To address this, R\textsuperscript{2}-Gaussian introduces a covariance-related factor to scale the density and achieve accurate density retrieval. Nevertheless, the projection process still relies on the local affine approximation:

\begin{equation}
\begin{aligned}
I_r(\mathbf{r}) &\approx \sum_{i=1}^{M} \int G_i \left( \tilde{\mathbf{x}} \mid \rho_i, \phi(\mathbf{p}), \mathbf{J}_i\mathbf{W}\mathbf{\Sigma}_i\mathbf{W}^\top\mathbf{J}_i^\top \right) \, dx_2,  \\
\end{aligned}
\end{equation}

\noindent where \(G_i\) denotes a 3D Gaussian kernel and \(\tilde{\mathbf{x}}=[x_0,x_1,x_2]\) represents a coordinate point in the ray space. In addition, \(\rho_i\), \(\boldsymbol{p}_i\), and \(\boldsymbol{\Sigma}_i\) are learnable parameters corresponding to central density, position and covariance, respectively. Moreover, \(\mathbf{J}_i\) is the local approximation matrix and \(\mathbf{W}\) is the viewing transformation matrix.

\begin{figure}[t]
  \centering
   \includegraphics[width=0.9\linewidth]{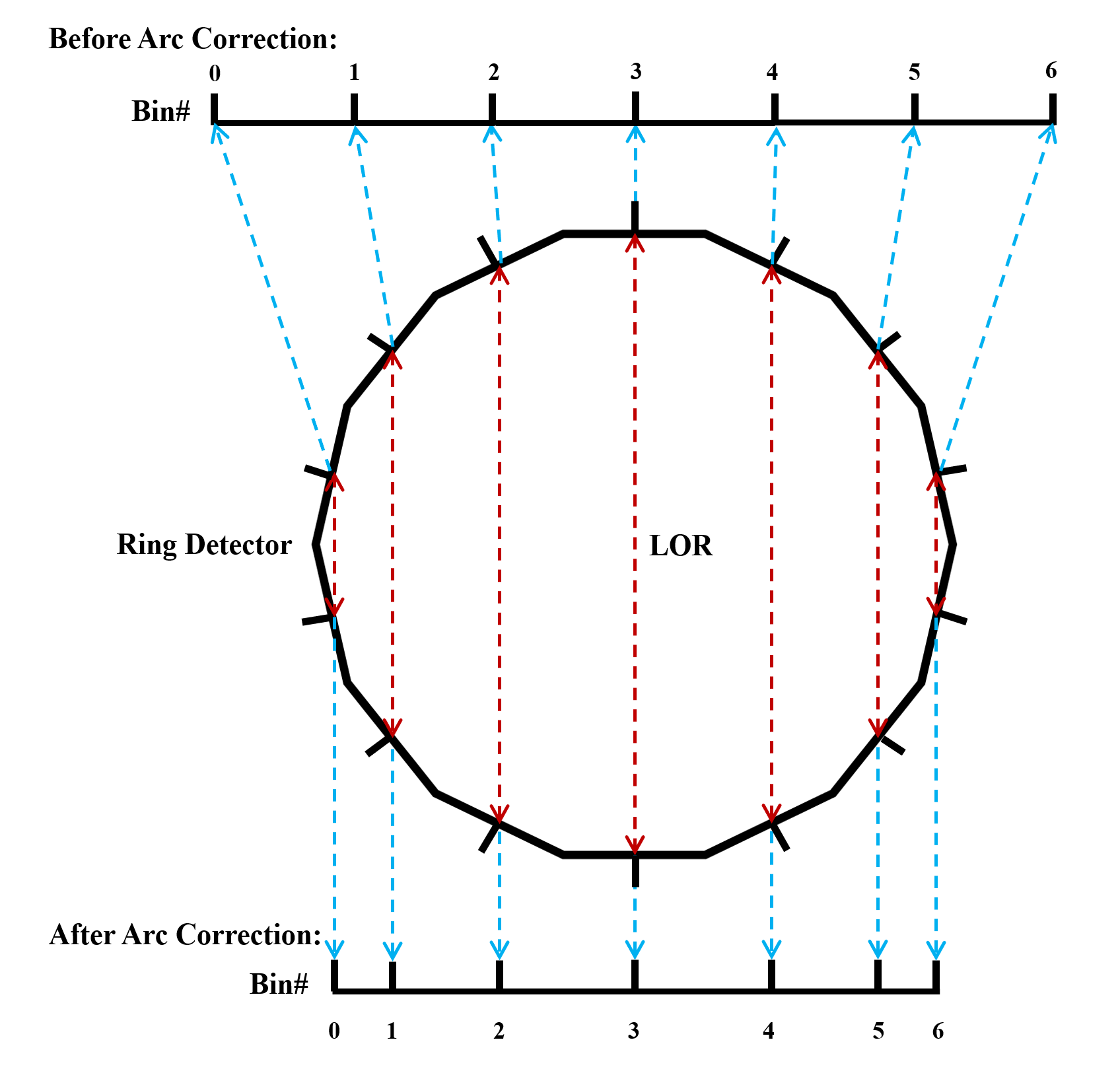}
   \caption{An example of a cylindrical detector arrangement that produces non-uniform tangential bin spacing in the sinogram. The sinogram before arc correction corresponds to straightening the arc-shaped detectors into a line (top one). However, the physical spacing between bins, namely, measured along chord, is non-uniform (bottom one).}
   \label{fig:arc_effect}
\end{figure}

\noindent\textbf{Inherent limitations.\quad} In applying R\textsuperscript{2}-Gaussian to PET image reconstruction, several challenges arise from the unique physical characteristics of the modality. R\textsuperscript{2}-Gaussian was originally designed for cone-beam or parallel-beam CT reconstruction. In PET, cross-segment LORs \cite{lewitt1994three,defrise1998data}, namely, those with non-zero ring difference, produce oblique projections, so we need to update the projector with a shear transformation; this update is straightforward to implement. A key consideration is the need for arc correction, which is required by the cylindrical arrangement of detectors in PET scanners. This geometry produces non-uniform bin spacing in the sinogram, as shown in \cref{fig:arc_effect}, requiring the forward projection model to accurately represent both the cylindrically arranged detectors and the varying spacings of LORs. Traditional 3DGS methods, which rely on local affine approximations, are difficult to adapt to such arc-corrected projections. The same limitation applies to the tomographic version of 3DGS, namely R\textsuperscript{2}-Gaussian. Although conservative resampling could be applied, it compromises the fidelity of the raw measured projections. Furthermore, PET is a quantitative imaging modality where reconstruction accuracy directly affects clinical measurements such as standardized uptake values (SUVs) \cite{thie2004understanding}, which are used to assess metabolic activity. Therefore, the projection model must maintain high quantitative accuracy and avoid approximations that could compromise diagnostic reliability or introduce bias in density estimation.

\subsection{Overall Framework}

These limitations motivate a tomographic reconstruction model that can not only provide more accurate line integrals but also be geometrically flexible for ray alignment. Thus, we employ 3D Gaussian ray tracing for tomographic reconstruction. By ray tracing, we can calculate the line integrals analytically without approximation and also define ray origins flexibly. Below we derive the analytic line integral of a 3D anisotropic Gaussian along a ray.

Let a single anisotropic 3D Gaussian, with the constant normalization absorbed into a density scalar, be written as:

\begin{equation}
G(\boldsymbol{x}) = \exp\!\Big(-\tfrac{1}{2}(\boldsymbol{x}-\boldsymbol{\mu})^\top \boldsymbol{\Sigma}^{-1}(\boldsymbol{x}-\boldsymbol{\mu})\Big),
\end{equation}

\noindent where \(\boldsymbol{x}\in\mathbb{R}^3\) represents a coordinate point in the ray space, \(\boldsymbol{\mu}\) is the mean that controls the position of the 3D Gaussian ellipsoid, and \(\boldsymbol{\Sigma}\) is the covariance, which can be decomposed into a scale matrix \(\boldsymbol{S}\) and a rotation matrix \(\boldsymbol{R}\): \(\boldsymbol{\Sigma}=\boldsymbol{R}\boldsymbol{S}\boldsymbol{S}^\top\boldsymbol{R}^\top\). Since a learnable density scalar \(\rho\), which can be adjusted during optimization, multiplies the 3D Gaussian kernel during the line integral, there is no need to explicitly handle the normalization term \(\frac{1}{(2\pi)^{3/2}|\boldsymbol{\Sigma}|^{1/2}}\).

A ray \(\boldsymbol{r}(t)\) is parameterized by the ray origin \(\boldsymbol{o}\) and the ray direction \(\boldsymbol{d}\):

\begin{equation}
\boldsymbol{r}(t) = \boldsymbol{o}+t\boldsymbol{d},
\end{equation}

\noindent where \(t\in\mathbb{R}\), \(\boldsymbol{o}\in\mathbb{R}^3\), and \(\boldsymbol{d}\in\mathbb{R}^3\). We consider the line integral over all 3D Gaussians as:
\begin{equation}
\begin{aligned}
I(\boldsymbol{r}) &= \int_{-\infty}^{\infty} \sum_{i=1}^{M} \rho_i \cdot G_i \left( \boldsymbol{r}(t)  \right) \, dt, \\
I(\boldsymbol{o}, \boldsymbol{d})     &= \sum_{i=1}^{M} \rho_i \cdot \int_{-\infty}^{\infty} G_i \left( \boldsymbol{o}+t\boldsymbol{d}  \right) \, dt,
\end{aligned}
\end{equation}

\noindent where \(I(\boldsymbol{r})\) represents the rendered pixel value along ray \(\boldsymbol{r}\), \(M\) is the number of 3D Gaussians in the ray space, and \(\rho_i\) is the central density of the \(i\)-th 3D Gaussian.

\subsection{Quadratic Exponent Along the Ray}

We consider the line integral for a single Gaussian without density as:
\begin{equation}
\label{eq:single_integral}
\dot{I}(\boldsymbol{o}, \boldsymbol{d}) = \int_{-\infty}^{\infty}
\exp\!\Big(-\tfrac{1}{2}(\boldsymbol{o}+t\boldsymbol{d}-\boldsymbol{\mu})^\top \boldsymbol{\Sigma}^{-1}(\boldsymbol{o}+t\boldsymbol{d}-\boldsymbol{\mu})\Big)
\, dt,
\end{equation}

Define \(\boldsymbol{\delta}=\boldsymbol{o}-\boldsymbol{\mu}\). Substituting into the exponent yields:

\begin{equation}
\begin{aligned}
\label{eq:exponent}
\text{exponent}
&=-\tfrac{1}{2}(\boldsymbol{o}+t\boldsymbol{d}-\boldsymbol{\mu})^\top \boldsymbol{\Sigma}^{-1}(\boldsymbol{o}+t\boldsymbol{d}-\boldsymbol{\mu})\\
&=-\tfrac{1}{2}(\boldsymbol{\delta}+t\boldsymbol{d})^\top \boldsymbol{\Sigma}^{-1}(\boldsymbol{\delta}+t\boldsymbol{d}) \\
&=-\tfrac{1}{2}(\boldsymbol{\delta}^\top\boldsymbol{\Sigma}^{-1}\boldsymbol{\delta} + 2t\,\boldsymbol{d}^\top\boldsymbol{\Sigma}^{-1}\boldsymbol{\delta} + t^2\,\boldsymbol{d}^\top\boldsymbol{\Sigma}^{-1}\boldsymbol{d}).
\end{aligned}
\end{equation}

Since \(\boldsymbol{\Sigma}^{-1}\) is symmetric, \(\boldsymbol{\delta}^\top\boldsymbol{\Sigma}^{-1}\boldsymbol{d} = \boldsymbol{d}^\top\boldsymbol{\Sigma}^{-1}\boldsymbol{\delta}\). Here, we define:

\begin{equation}
    \begin{aligned}
    \label{eq:ABC}
        A &= \boldsymbol{d}^\top \boldsymbol{\Sigma}^{-1} \boldsymbol{d}, \\
        B &= \boldsymbol{d}^\top \boldsymbol{\Sigma}^{-1} \boldsymbol{\delta}, \\
        C &= \boldsymbol{\delta}^\top \boldsymbol{\Sigma}^{-1} \boldsymbol{\delta}.
    \end{aligned}
\end{equation}

Substituting \cref{eq:ABC} into \cref{eq:exponent} gives the quadratic form of the exponent along the ray:

\begin{equation}
\label{eq:quadratic_exponent}
\text{exponent} = -\tfrac{1}{2}(C + 2Bt +At^2).
\end{equation}

\subsection{Closed Form Integral}

Substituting the quadratic expression into \cref{eq:single_integral}, the integral becomes:
\begin{equation}
\label{eq:single_ABC_integral}
\begin{aligned}
\dot{I}(\boldsymbol{o}, \boldsymbol{d})
&=  \int_{-\infty}^{\infty} \exp\!\Big( -\tfrac{1}{2}(C + 2Bt +At^2) \Big) \, dt, \\
&= \exp\!\Big( -\tfrac{1}{2}C \Big) \int_{-\infty}^{\infty} \exp\!\Big( -\tfrac{1}{2}(2Bt +At^2) \Big) \, dt, \\
&= \exp\!\Big( -\tfrac{1}{2}C \Big) \int_{-\infty}^{\infty} \exp\!\Big( -\tfrac{1}{2}A(t + \tfrac{B}{A})^2 + \tfrac{B^2}{2A} \Big) \, dt, \\
&= \exp\!\Big( -\tfrac{1}{2}(C - \tfrac{B^2}{A}) \Big)  \int_{-\infty}^{\infty} \exp\!\Big(
-\tfrac{1}{2}A(t + \tfrac{B}{A})^2 \Big) \, dt.
\end{aligned}
\end{equation}

Let \(h=t+\tfrac{B}{A}\). The remaining integral is the standard Gaussian integral:
\begin{equation}
 \int_{-\infty}^{\infty} \exp\!\Big( -\tfrac{1}{2}Ah^2 \Big) \, dh = \sqrt{\tfrac{2\pi}{A}}.
\end{equation}

Therefore, the line integral for a single Gaussian without density is:

\begin{equation}
\dot{I}(\boldsymbol{o}, \boldsymbol{d}) = \sqrt{\tfrac{2\pi}{A}} \cdot \exp\!\Big( -\tfrac{1}{2}\big(C - \tfrac{B^2}{A}\big) \Big).
\end{equation}

Finally, the line integral along the ray for all Gaussians becomes:

\begin{equation}
\boxed{I(\boldsymbol{r}) =
\sum_{i=1}^{M} \rho_i \cdot \sqrt{\tfrac{2\pi}{A_i}} \cdot \exp\!\Big( -\tfrac{1}{2}\big(C_i - \tfrac{B_i^2}{A_i}\big) \Big)}
\end{equation}

This expression gives an analytic, differentiable forward projection for the 3D Gaussian model and avoids the local affine projection approximation commonly used in splatting-based methods. 
\section{Experiment}
\label{sec:Experiment}

\subsection{Experimental Settings}

\noindent\textbf{PET dataset.\quad} The PET dataset contains three parts. First, we analytically simulate the NEMA phantom using the Software for Tomographic Image Reconstruction (STIR) v6.2 \cite{thielemans2012stir} for a Siemens Biograph mMR scanner and add Poisson noise to the sinogram afterward. STIR can simulate arc-corrected sinograms by resampling the raw projections using an overlap-interpolation method. Then, for the three-point-source Monte Carlo simulation to investigate the influence of arc versus non-arc correction, we follow the protocol in the PET/CT Acceptance Testing and Quality Assurance Report from the American Association of Physicists in Medicine (AAPM) \cite{lopez2021pet}, using the GEANT4 Application for Tomographic Emission (GATE) v9.2 \cite{strulab2003gate}. In the simulation, three point sources are located at (0, 1), (0, 10), and (0, 20) cm, respectively, where (0,0) cm is the isocenter of the scanner. The stopping condition for the simulation is acquisition of 5 million counts. The scanner used in this simulation was a General Electric Discovery 690. The raw output data are converted to sinograms for reconstruction by STIR. Finally, for qualitative assessment, a realistic brain data was collected from a collaborating institute, which was used under approval by the institutional review board, and informed consent was waived because the data were analyzed retrospectively and anonymized.

\noindent\textbf{CT dataset.\quad} To facilitate comparison with R\textsuperscript{2}-Gaussian, we employ the same synthetic and real CT datasets to evaluate performance differences. To establish a benchmark for sparse-view reconstruction, three scenarios with 75, 50, and 25 views are defined and applied to both synthetic and real-world data. The synthetic data were derived from 15 diverse CT volumes of natural and artificial objects, including human organs, animals and plants, and artificial objects. X-ray projections were computationally generated using the TIGRE toolbox \cite{biguri2016tigre} and subsequently contaminated with models of electronic noise. For real-world data evaluation, three objects from the Finnish Inverse Problems Society (FIPS) dataset are utilized, each containing 721 real projections. In the absence of definitive ground truths, pseudo-ground-truth volumes were generated by applying the FDK algorithm \cite{feldkamp1984practical} to the complete set of views.

\noindent\textbf{Implementation details.\quad} The implementation of 3D Gaussian ray tracing for tomographic reconstruction is based on PyTorch \cite{paszke2019pytorch} and the NVIDIA OptiX ray tracing engine v6.0 \cite{parker2010optix}. All experiments are run using the Adam optimizer \cite{adam2014method} with the same Gaussian densify-and-prune strategy as R\textsuperscript{2}-Gaussian \cite{zha2024r}. For ray tracing, an icosahedron is used to represent a Gaussian. Back-face culling is enabled to trigger our custom Anyhit program for accumulating the analytic integral of each Gaussian. To accelerate the training process, we also skip Gaussians whose contributions are negligible. All tasks are conducted on an RTX 3080 Ti GPU.

\subsection{Experimental Results}

\begin{table}
  \caption{Comparison of reconstruction times for the NEMA phantom.}
  \label{tab:recon_time}
  \centering
  \small
  \setlength{\tabcolsep}{3pt}
  \begin{tabular}{@{}lccc@{}}
    \toprule
    Method & OSEM & R\textsuperscript{2}-Gaussian & Our method \\
    \midrule
    Time (min)& 17 & \textbf{12} & 35 \\
    \bottomrule
  \end{tabular}
\end{table}

The NEMA phantom contains six spheres of different sizes with a hollow cylinder in the middle. The six spheres are numbered 0 to 5 according to their inner diameter, which are 10 mm, 13 mm, 17 mm, 22 mm, 28 mm, and 37 mm, respectively. Since R\textsuperscript{2}-Gaussian previously only worked for parallel-beam or cone-beam projections, we updated the projector to make it feasible for oblique projections, which is necessary for utilizing cross-segment LORs in PET reconstruction. Furthermore, we compare with the OSEM implementation in STIR as a baseline. We compare three reconstruction results: (1) noisy NEMA phantom reconstructed by OSEM; (2) noisy NEMA phantom reconstructed by R\textsuperscript{2}-Gaussian; and (3) noisy NEMA phantom reconstructed by our method. Reconstruction performance is evaluated in terms of (i) signal-to-background ratio (SBR), (ii) recovered sphere diameters, and (iii) the standard deviation within each sphere. These metrics indicate quantitative accuracy, reconstruction fidelity, and noise level, respectively. The reconstruction times are summarized in \cref{tab:recon_time}. Owing to the highly parallel nature of splatting-based methods, R\textsuperscript{2}-Gaussian is the fastest of the three methods. OSEM is intermediate. Our method is the slowest, taking almost twice as long as R\textsuperscript{2}-Gaussian.

\begin{table*}
  \caption{Measured diameters (mm) for each NEMA phantom sphere reconstructed by OSEM, R\textsuperscript{2}-Gaussian, and our method. Parentheses show relative error versus ground truth.}
  \label{tab:nema_diam}
  \centering
  \small
  \setlength{\tabcolsep}{3pt}
  \begin{tabular}{@{}lcccccc@{}}
    \toprule
    Method                        & Sphere0       & Sphere1        & Sphere2        & Sphere3       & Sphere4        & Sphere5 \\
    \midrule
    OSEM                          & 10.9 (8.9\%)  & 13.7 (-21.4\%) & \textbf{17.2 (1.2\%)}   & 20.6 (-6.3\%) & \textbf{27.7 (-1.0\%)}  & \textbf{37.3 (0.7\%)} \\
    R\textsuperscript{2}-Gaussian & \textbf{10.0 (-0.4\%)} & 10.8 (17.1\%)  & 15.7 (-7.4\%)  & \textbf{21.3 (-3.4\%)} & \textbf{28.4 (1.3\%)}   & 39.2 (6.0\%)\\
    Our method                    & 9.5 (-5.1\%)  & 10.8 (16.6\%)  & 15.1 (-11.1\%) & \textbf{21.2 (-3.7\%)} & \textbf{27.6 (-1.3\%)}  & 39.7 (7.2\%)\\
    \bottomrule
  \end{tabular}
\end{table*}

\cref{tab:nema_diam} summarizes the reconstructed sphere diameters for the three methods. The diameter is obtained by the full width half maximum of a sphere. From this table, both R\textsuperscript{2}-Gaussian and OSEM have three out of six spheres with diameter error less than 5\%. Notably, R\textsuperscript{2}-Gaussian perfectly reconstructed the smallest sphere (10 mm), which is generally the most challenging target. Our method has two out of six spheres with error less than 5\%, suggesting that recovery of true diameters by our method is slightly inferior to R\textsuperscript{2}-Gaussian and OSEM in these experiments.

\begin{table*}
  \caption{Signal-to-background ratios for each NEMA phantom sphere reconstructed by OSEM, R\textsuperscript{2}-Gaussian, and our method. Parentheses show percentage error relative to the reference activity ratio (4:1).}
  \label{tab:nema_sbr}
  \small
  \setlength{\tabcolsep}{3pt}
  \centering
  \begin{tabular}{@{}lcccccc@{}}
    \toprule
    Method                        & Sphere0        & Sphere1        & Sphere2        & Sphere3        & Sphere4       & Sphere5 \\
    \midrule
    OSEM                          & 2.15 (-46.3\%)          & 3.14 (-21.4\%)           & 3.35 (-16.1\%)           & 3.75 (-6.2\%)            & 3.78 (-5.4\%)            &  \textbf{3.99 (-0.2\%)}\\
    R\textsuperscript{2}-Gaussian & 2.99 (-25.2\%)          & 4.38 (9.6\%)             &  \textbf{4.14 (3.4\%)}   & 4.24 (6.1\%)             & 4.20 (5.1\%)             &  \textbf{4.06 (1.5\%)}\\
    Our method                    & 3.00 (-24.5\%) &  \textbf{3.97 (-0.7\%)}  &  \textbf{3.90 (-2.6\%)}  &  \textbf{4.16 (3.9\%)}   &  \textbf{4.13 (3.2\%)}  &  \textbf{4.07 (1.8\%)}\\
    \bottomrule
  \end{tabular}
\end{table*}

The activity ratio between spheres and background is set to 4:1; an SBR closer to 4 indicates better quantitative accuracy. \cref{tab:nema_sbr} summarizes the SBR results for all spheres.To calculate the SBR, we use the measured FWHM as the sphere diameter, compute the mean activity within that diameter, and obtain the SBR by dividing this mean by the mean activity of a 20×20×20 background volume.From the table, our method provides the best quantitative accuracy among the three methods: five out of six spheres have relative error smaller than 5\%. R\textsuperscript{2}-Gaussian is the second best, with two out of six spheres within 5\% error. OSEM performs worst in SBR accuracy, with only one sphere within 5\% error.

\begin{figure}[t]
  \centering
   \includegraphics[width=0.9\linewidth]{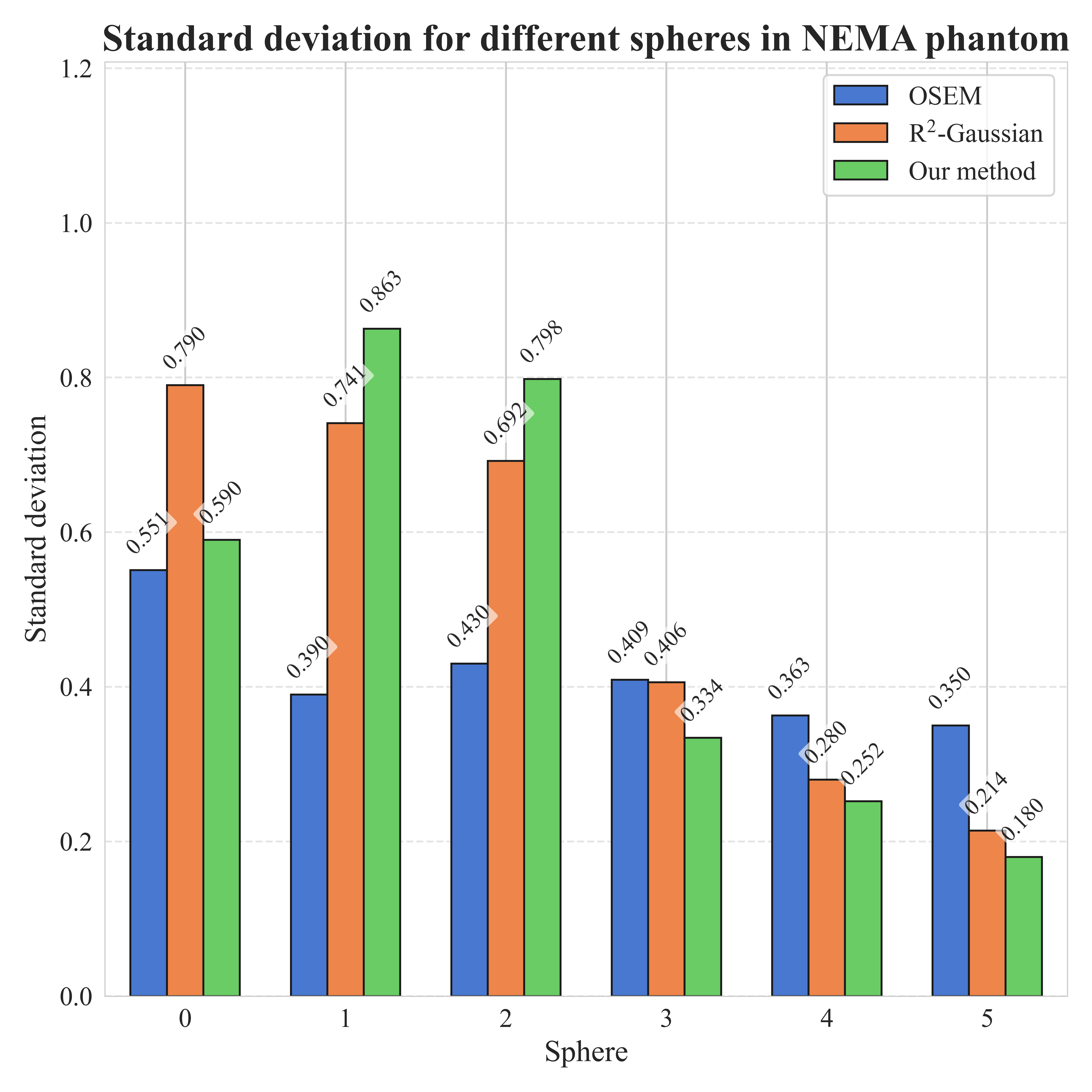}
   \caption{Standard deviation within each NEMA phantom sphere for OSEM, R\textsuperscript{2}-Gaussian, and our method (lower values indicate better noise suppression).}
   \label{fig:nema_std_bar}
\end{figure}

\cref{fig:nema_std_bar} shows the standard deviation inside each sphere for the three methods to quantify noise level. OSEM exhibits the lowest standard deviations overall, indicating stronger noise suppression. Both our method and R\textsuperscript{2}-Gaussian show higher standard deviations, particularly for the smaller spheres, which indicates higher noise levels in those regions. Taken together, the results indicate a trade-off: our method achieves the best quantitative accuracy across most spheres, while OSEM yields the lowest noise, and R\textsuperscript{2}-Gaussian provides a balance with good fidelity for small objects in some cases.

\begin{figure}[t]
  \centering
  \begin{subfigure}{0.8\linewidth}
    \centering
    \includegraphics[width=0.49\linewidth]{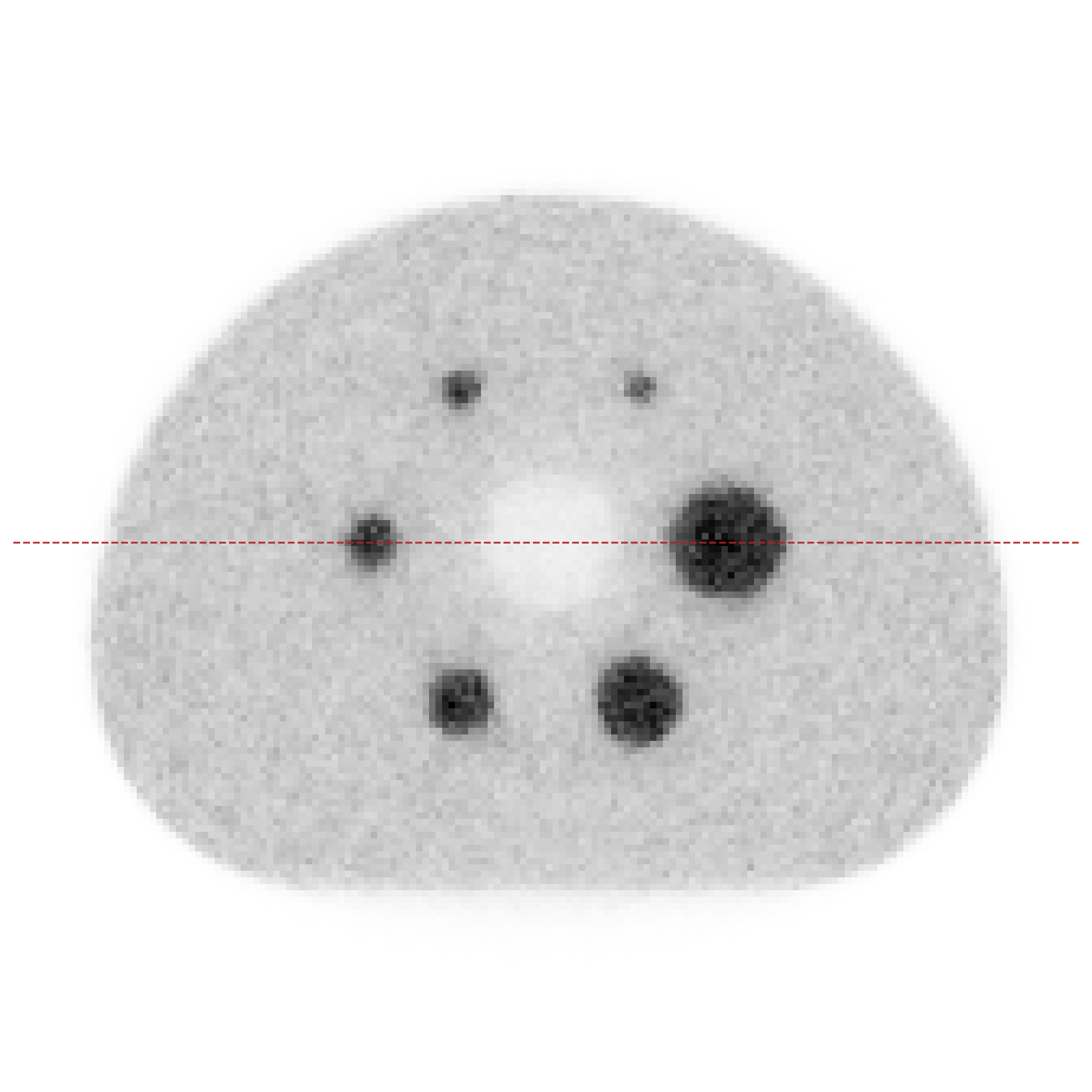}
    \hfill
    \includegraphics[width=0.49\linewidth]{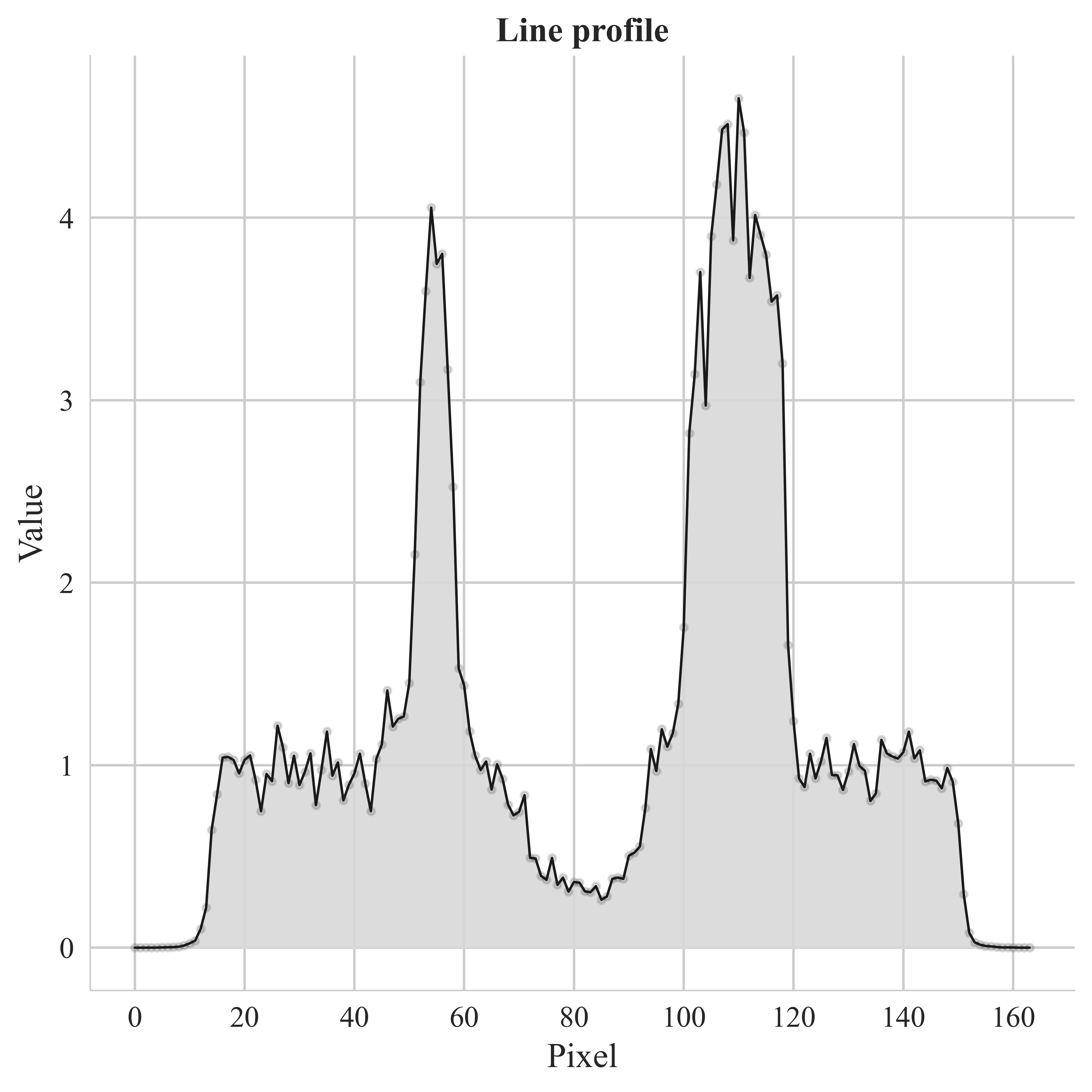}
    \caption{OSEM.}
    \label{fig:nema_line_profile_osem}
  \end{subfigure}
  \vskip\baselineskip
  \begin{subfigure}{0.8\linewidth}
    \centering
    \includegraphics[width=0.49\linewidth]{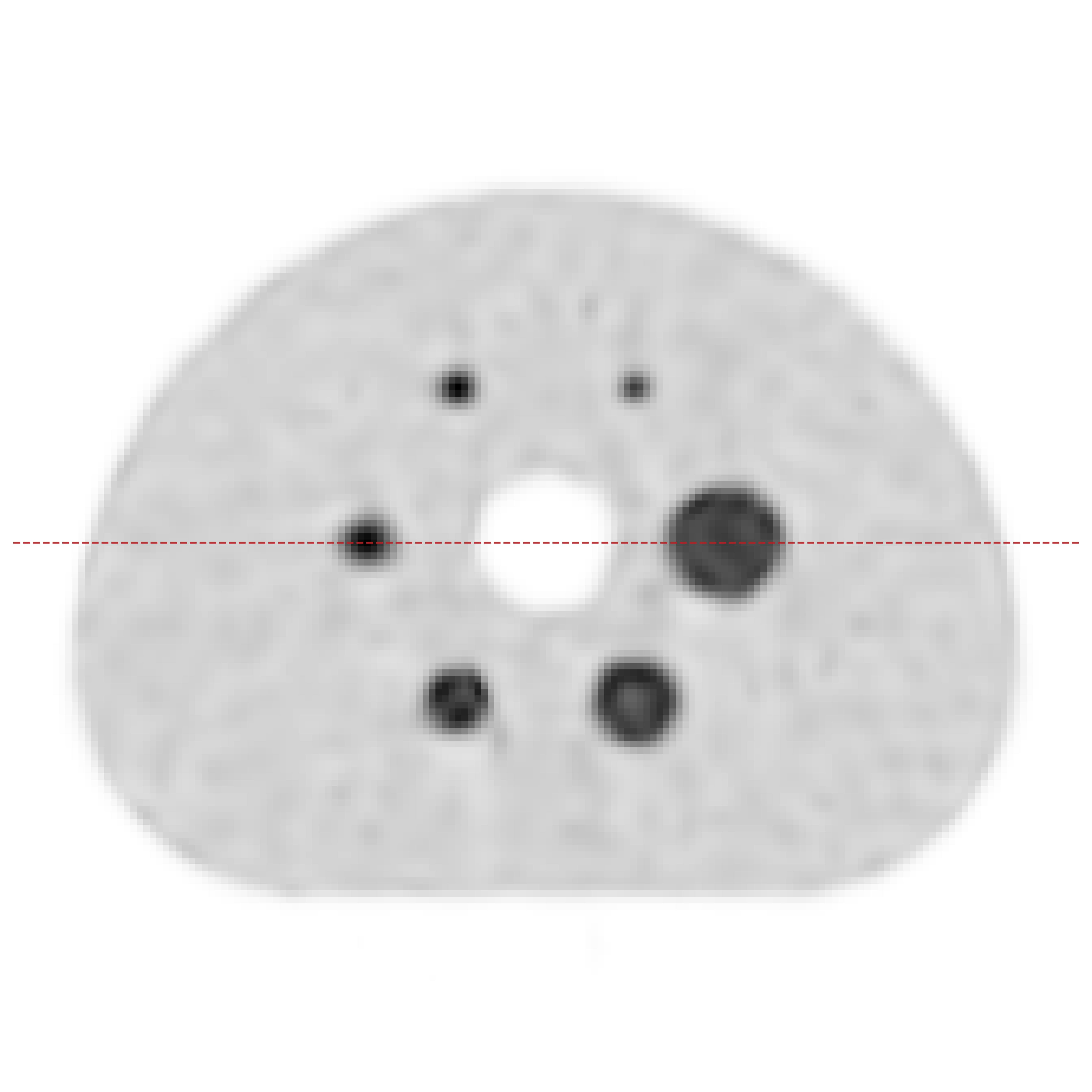}
    \hfill
    \includegraphics[width=0.49\linewidth]{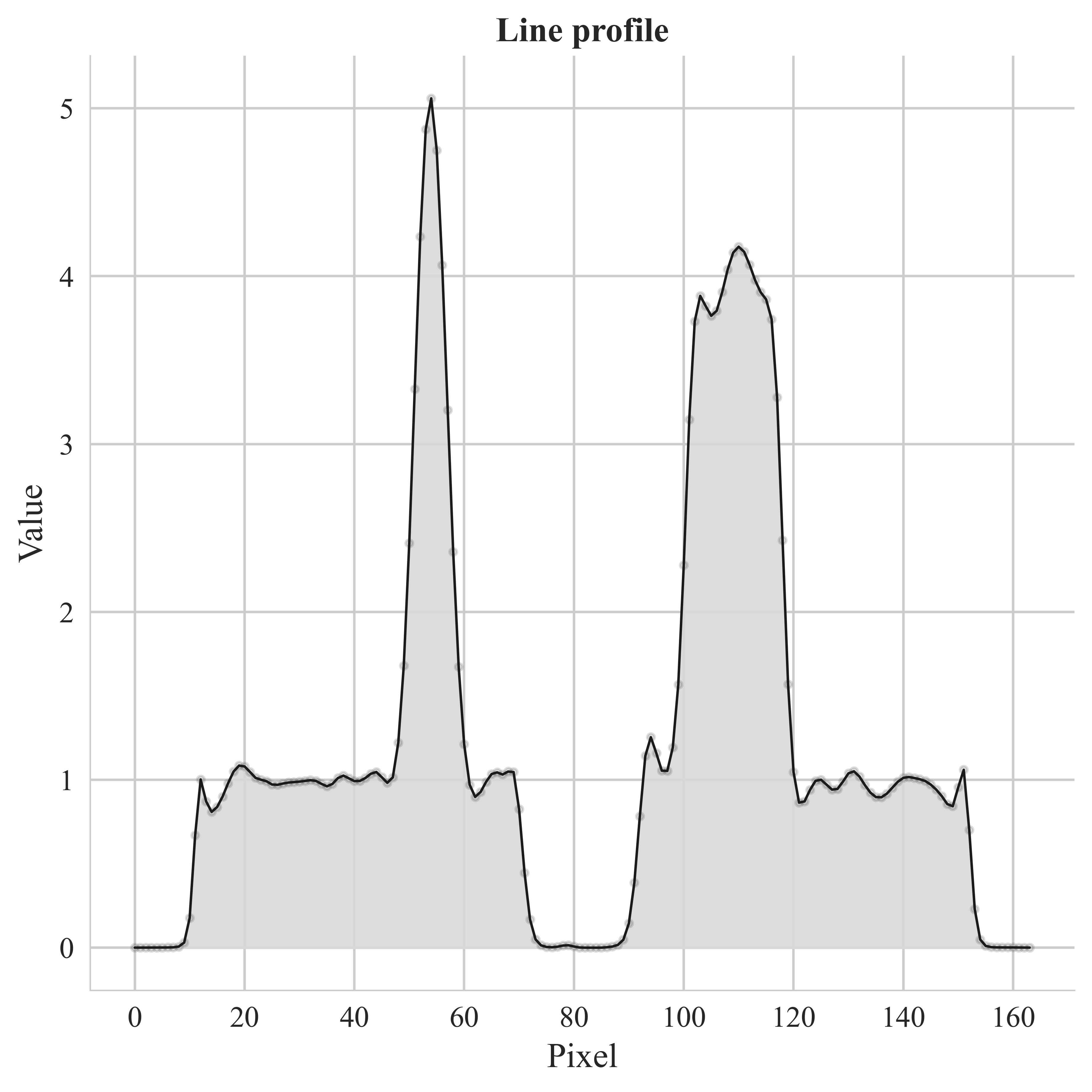}
    \caption{R\textsuperscript{2}-Gaussian.}
    \label{fig:nema_line_profile_r2}
  \end{subfigure}
  \vskip\baselineskip
  \begin{subfigure}{0.8\linewidth}
    \centering
    \includegraphics[width=0.49\linewidth]{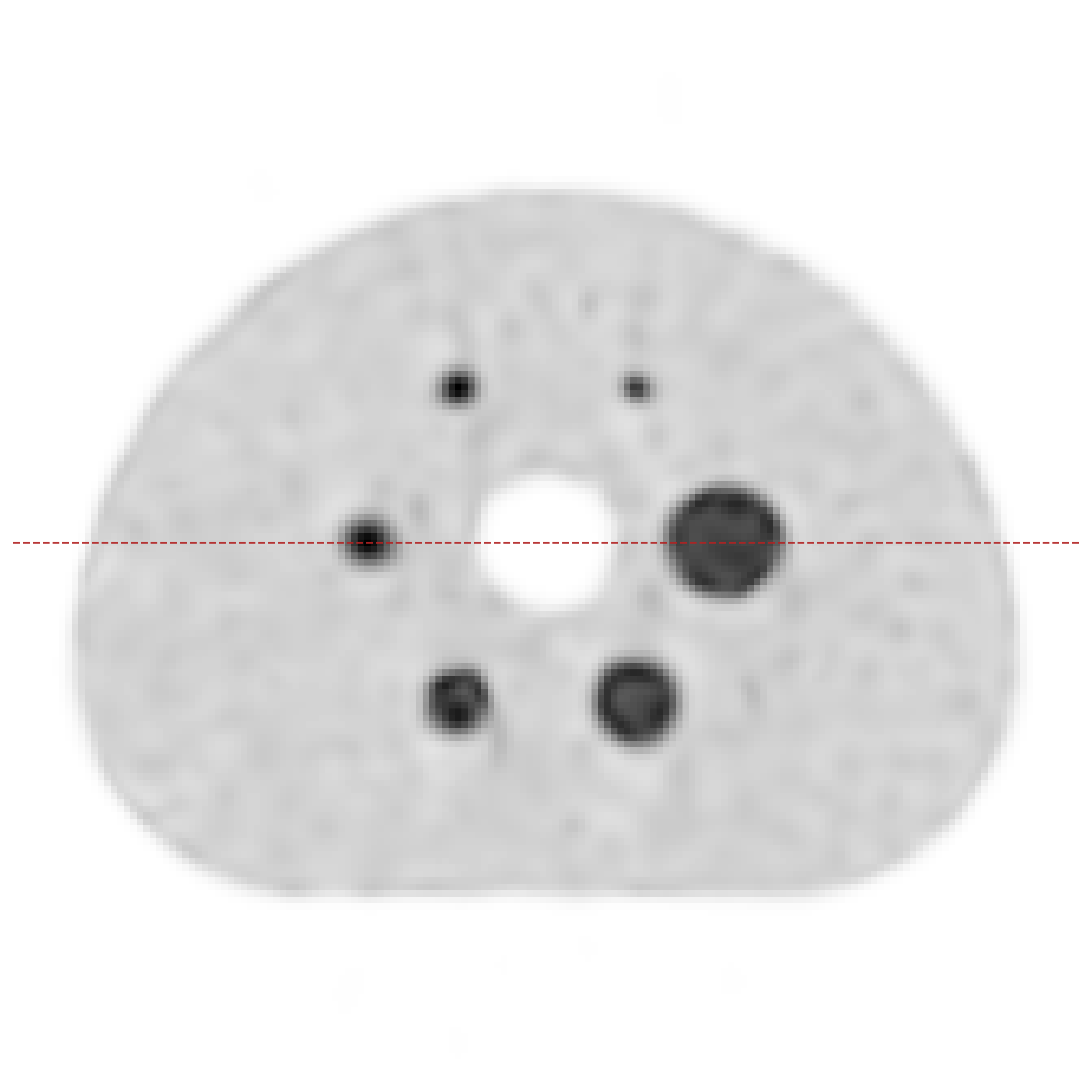}
    \hfill
    \includegraphics[width=0.49\linewidth]{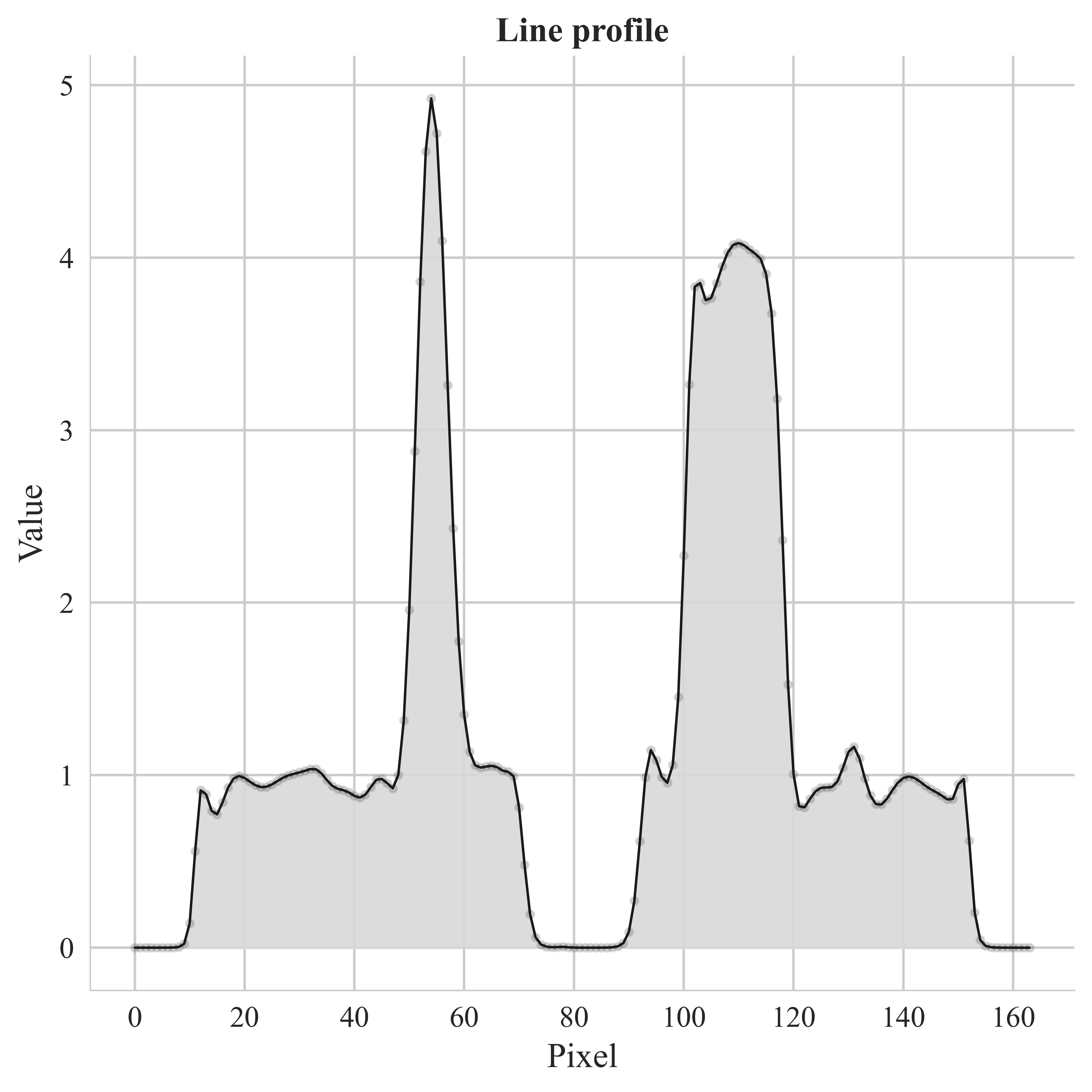}
    \caption{Our method.}
    \label{fig:nema_line_profile_3dgrt}
  \end{subfigure}
  \caption{NEMA phantom PET reconstructions by (a) OSEM, (b) R\textsuperscript{2}-Gaussian, and (c) our method. Corresponding central horizontal line profiles are plotted on the right.}
  \label{fig:nema_line_profile}
\end{figure}

\cref{fig:nema_line_profile} plots the line profile along the central horizontal line of the NEMA phantom reconstructions produced by each method. Note that the central cylinder of the NEMA phantom is empty and therefore should have zero activity. From the line profiles on the right, we can directly compare how the three methods preserve the zero background in the hollow cylinder: OSEM shows a positive bias in reconstructed activity inside the hollow cylinder, as shown in \cref{fig:nema_line_profile_osem}, indicating contamination or a background offset; R\textsuperscript{2}-Gaussian and our method stay close to zero there, showing they better preserve the low-signal property of the hollow region. The positive bias in OSEM may come from incomplete correction for scatter and randoms. However, methods based on 3D Gaussians can better suppress such biases.

\begin{figure}[t]
  \centering
  \begin{subfigure}{0.8\linewidth}
    \centering
    \includegraphics[width=0.49\linewidth]{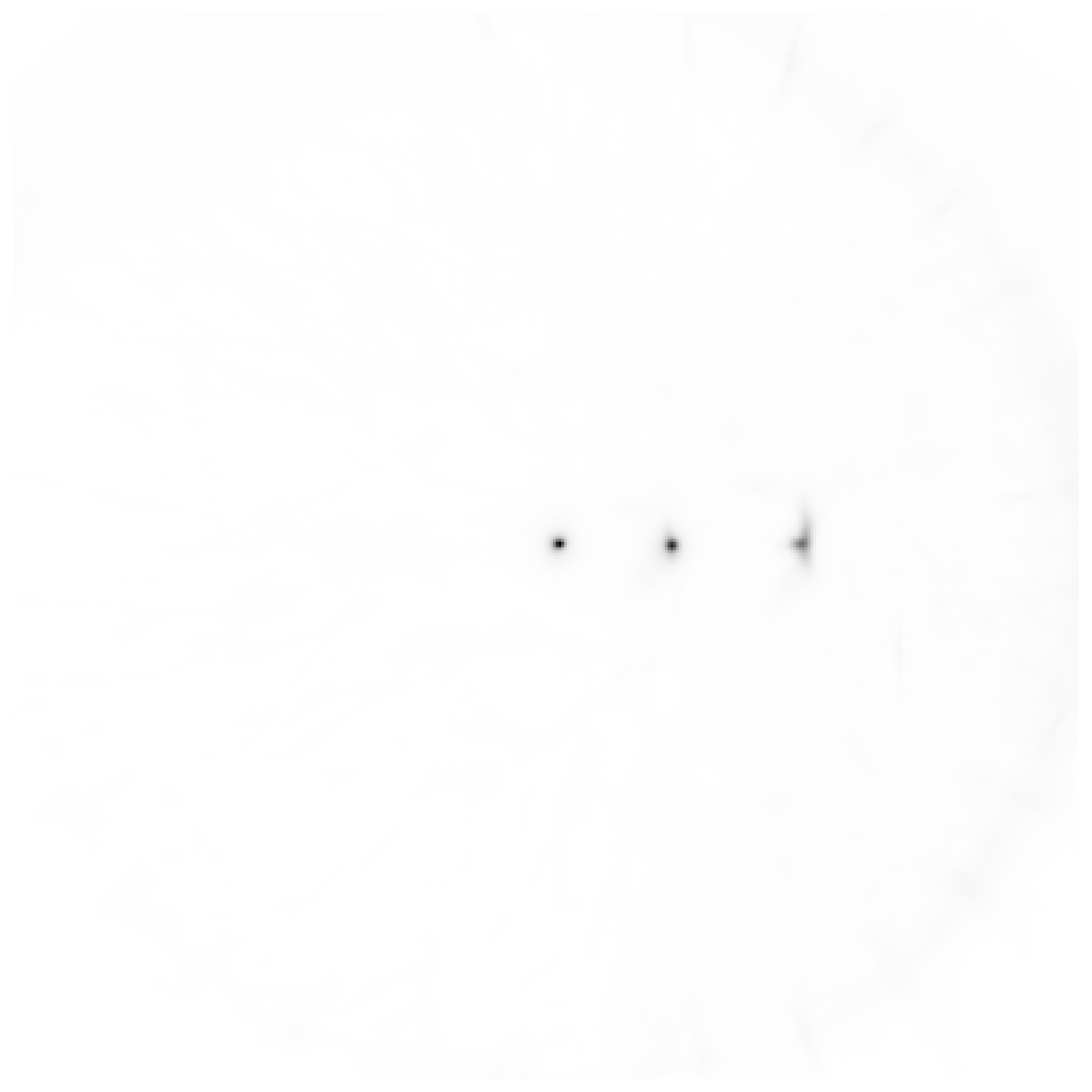}
    \hfill
    \includegraphics[width=0.49\linewidth]{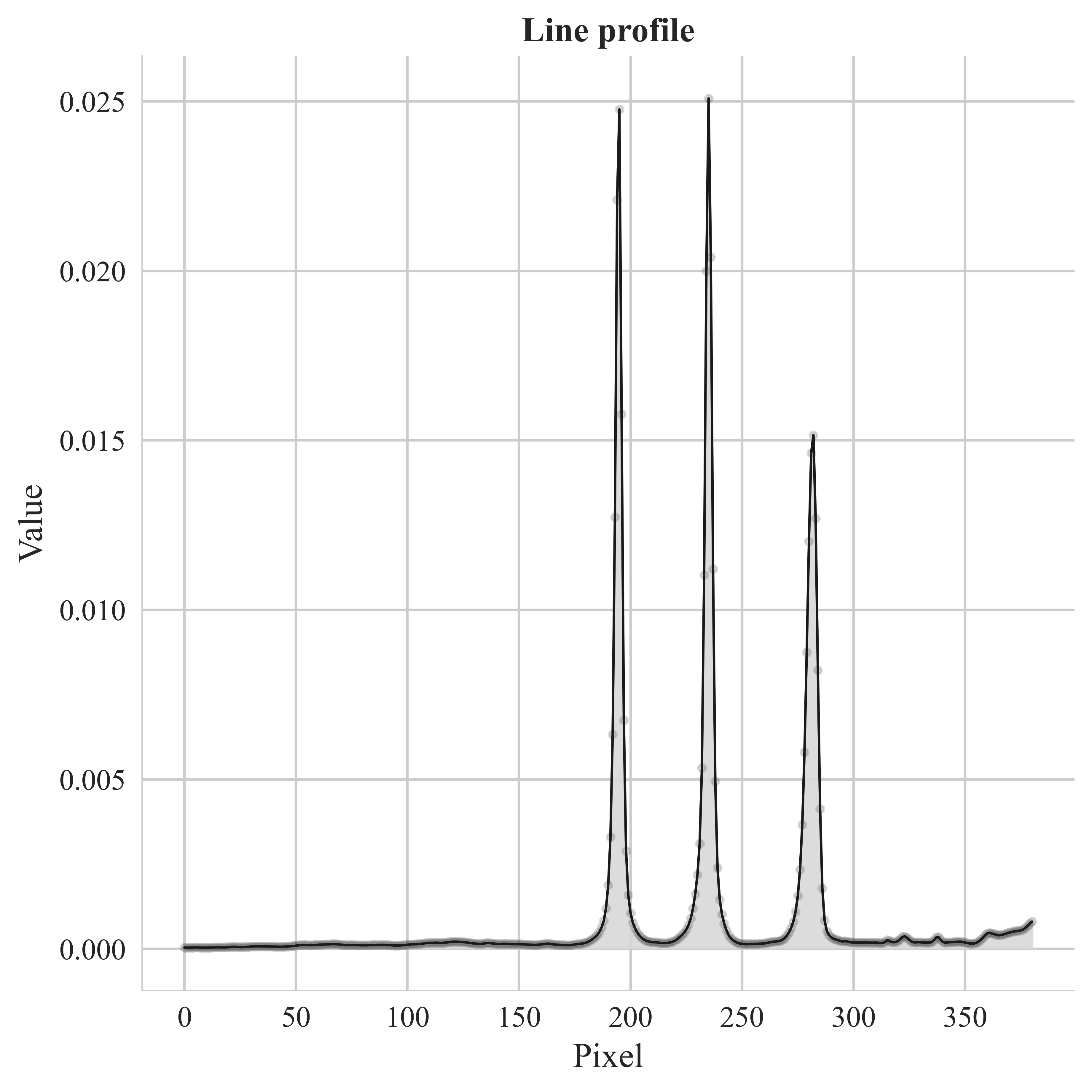}
    \caption{Non-arc-correction.}
    \label{fig:non-arc}
  \end{subfigure}
  \vskip\baselineskip
  \begin{subfigure}{0.8\linewidth}
    \centering
    \includegraphics[width=0.49\linewidth]{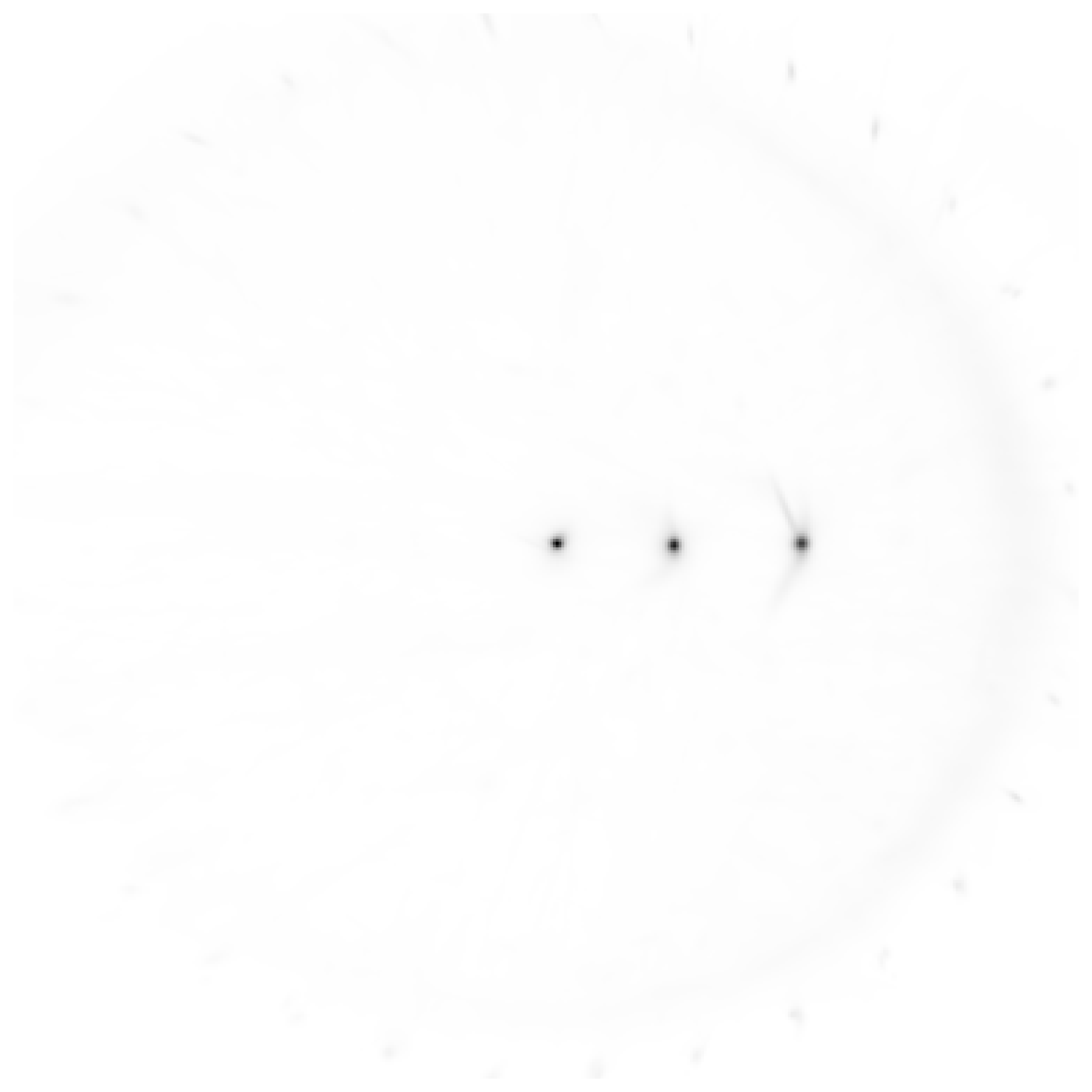}
    \hfill
    \includegraphics[width=0.49\linewidth]{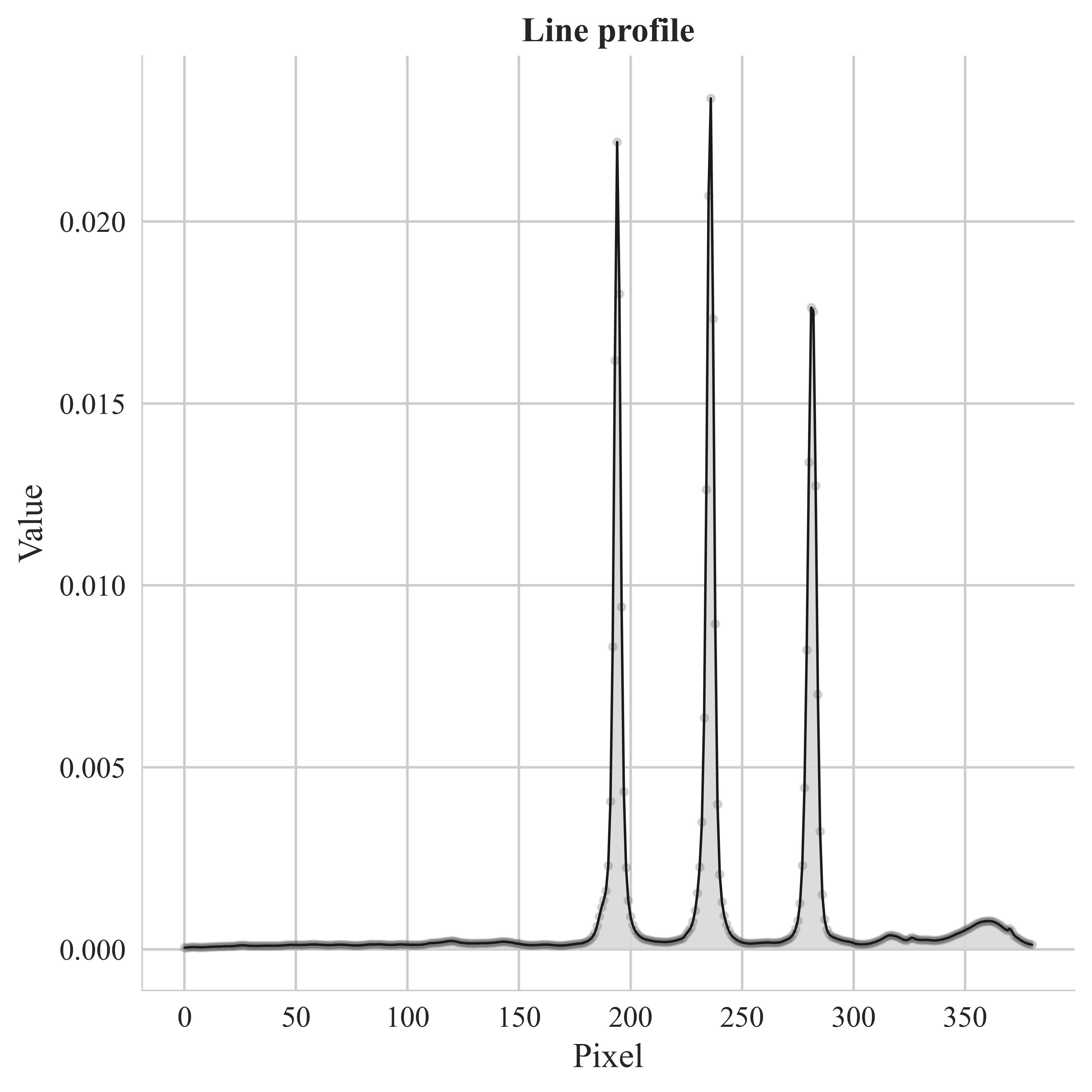}
    \caption{Arc correction.}
    \label{fig:arc}
  \end{subfigure}
    \caption{Three-point-source PET reconstructions: (a) without arc correction; (b) with arc correction. From left to right, sources are at (0,1), (0,10), and (0,20) cm. Central horizontal line profiles are shown on the right.}
  \label{fig:point_source}
\end{figure}

Point sources are commonly used to measure the spatial resolution of PET systems \cite{gong2016assessment}, since spatial resolution is typically defined as the FWHM of the Point Spread Function (PSF) \cite{akamatsu2014influences} and is calculated from the line profile through a reconstructed image of a point source in air. The results of the three-point-source Monte Carlo simulation are shown in \cref{fig:point_source}. All reconstructions were produced by our method; \cref{fig:non-arc} and \cref{fig:arc} show results without and with arc correction, respectively. Here we also measured the FWHM of the three point sources. For the non-arc case, the FWHMs are 3.43, 3.69, and 5.56 pixel, respectively. For the arc-corrected case, they are 3.45, 3.82, and 4.56 pixel, respectively. Since the three point sources were set to the same activity, their reconstructed values should be similar. The arc-corrected reconstruction yields values that are closer to each other. The results show that applying arc correction can improve spatial resolution and quantitative accuracy \cite{buchert2000performance, kadrmas2004lor}, especially for sources far from the center of field of view.

\begin{figure}[t]
  \centering
  \begin{subfigure}{0.8\linewidth}
    \centering
    \includegraphics[width=0.32\linewidth]{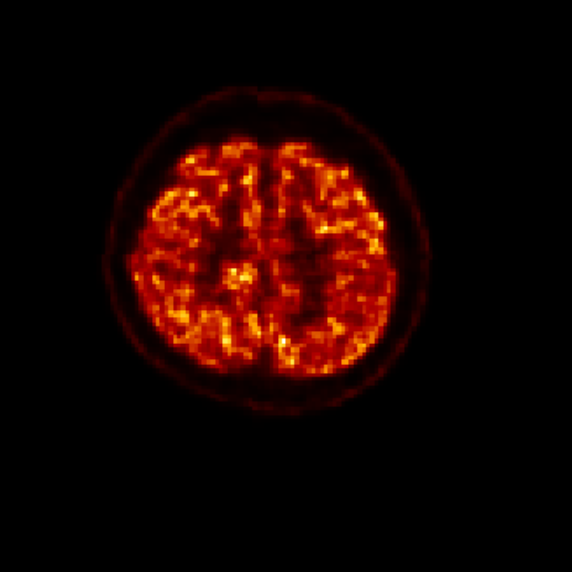}
    \hfill
    \includegraphics[width=0.32\linewidth]{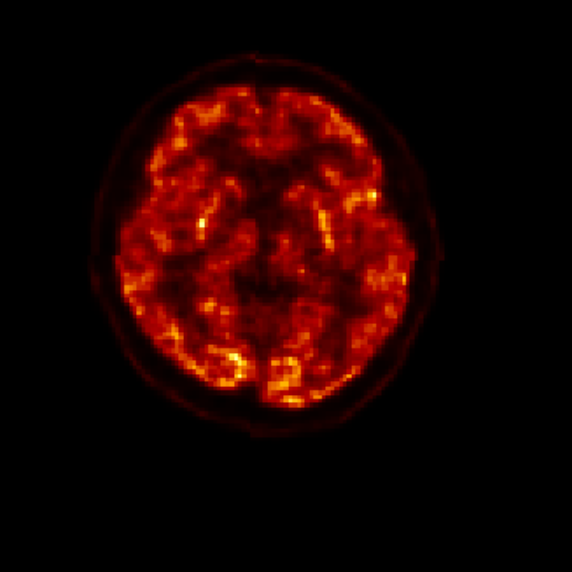}
    \hfill
    \includegraphics[width=0.32\linewidth]{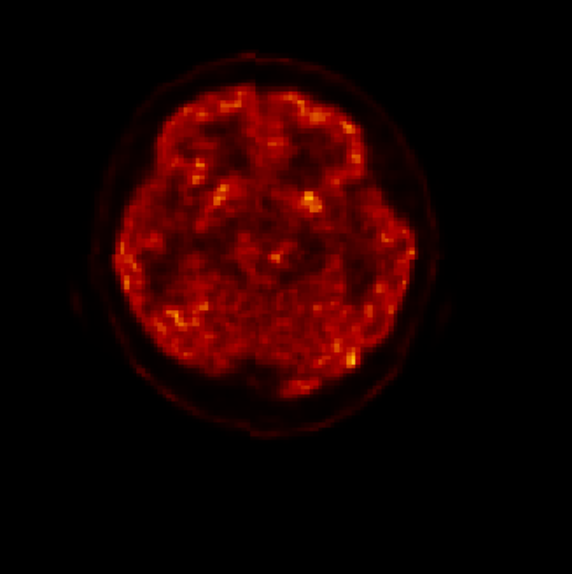}
    \caption{OSEM.}
    \label{fig:brain_OSEM}
  \end{subfigure}
  \vskip\baselineskip
  \begin{subfigure}{0.8\linewidth}
    \centering
    \includegraphics[width=0.32\linewidth]{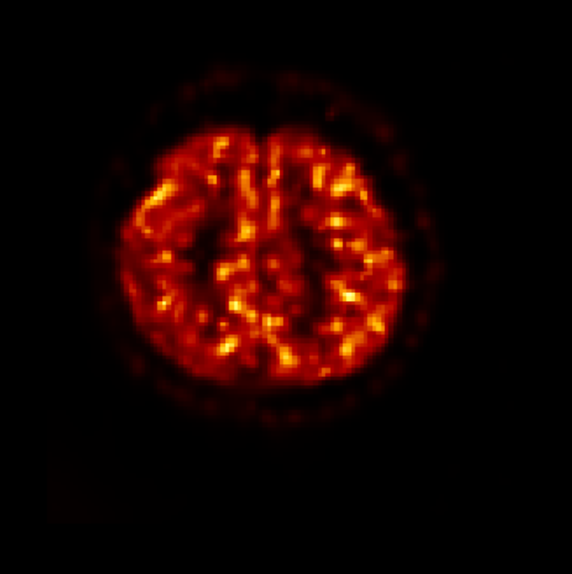}
    \hfill
    \includegraphics[width=0.32\linewidth]{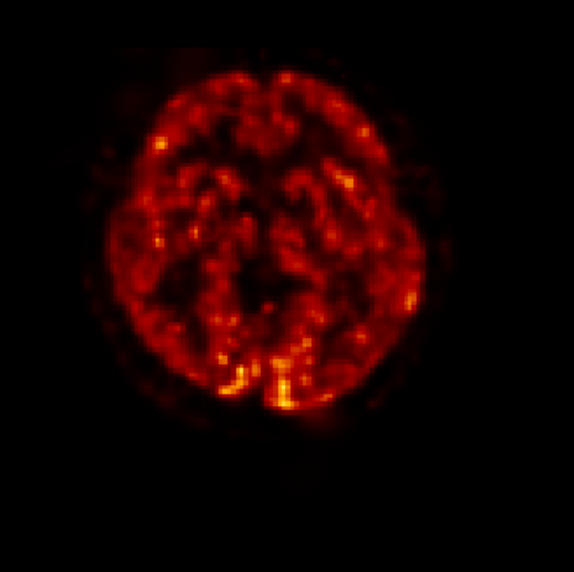}
    \hfill
    \includegraphics[width=0.32\linewidth]{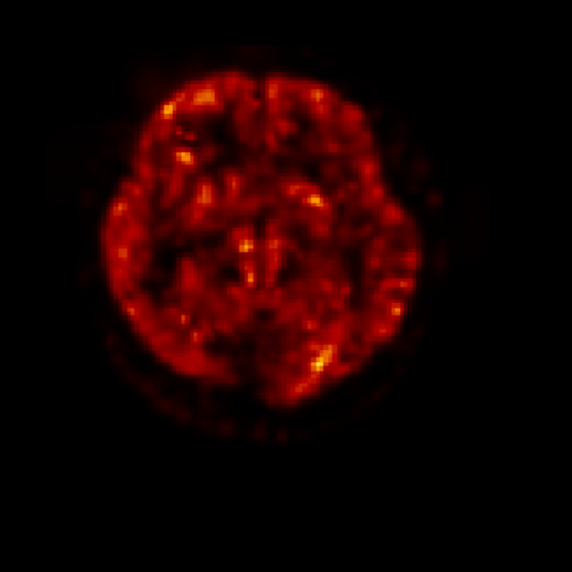}
    \caption{R\textsuperscript{2}-Gaussian.}
    \label{fig:brain_r2}
  \end{subfigure}
  \vskip\baselineskip
  \begin{subfigure}{0.8\linewidth}
    \centering
    \includegraphics[width=0.32\linewidth]{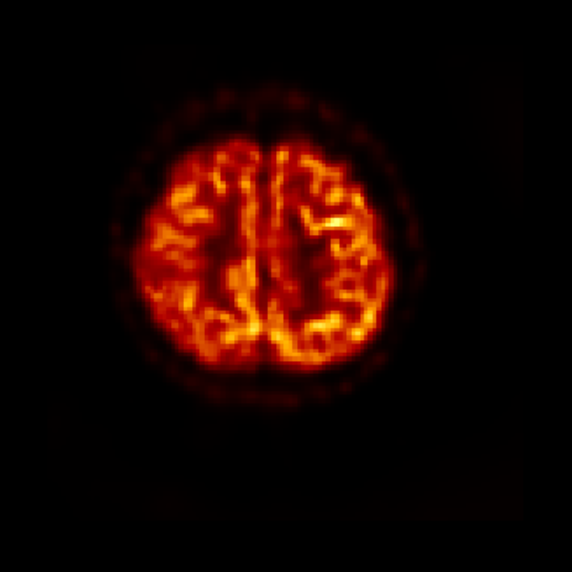}
    \hfill
    \includegraphics[width=0.32\linewidth]{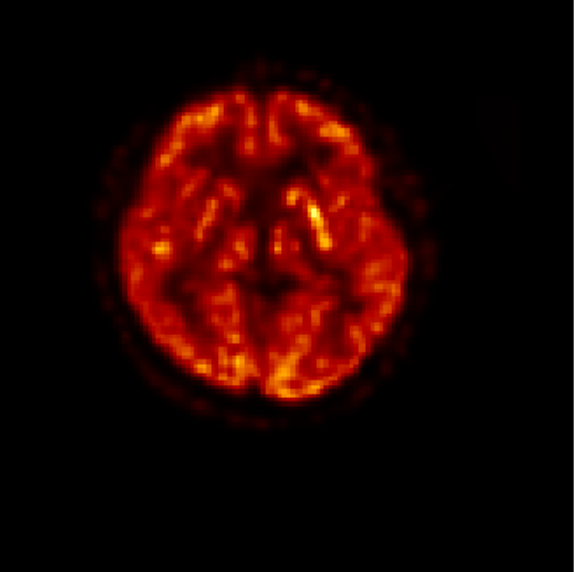}
    \hfill
    \includegraphics[width=0.32\linewidth]{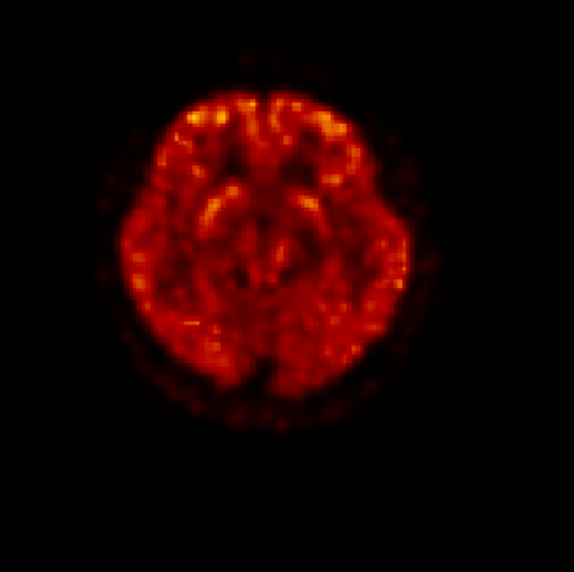}
    \caption{Our method.}
    \label{fig:brain_3dgrt}
  \end{subfigure}
  \caption{A realistic brain PET data reconstructions by (a) OSEM, (b) R\textsuperscript{2}-Gaussian, and (c) our method.}
  \label{fig:brain_data}
\end{figure}

\cref{fig:brain_data} shows reconstructions of realistic brain data by (a) OSEM, (b) R\textsuperscript{2}-Gaussian, and (c) our method. Visually, our method provides the most detail among the three. In the left column, the sulci and gyri of the cerebral cortex are more clearly visible in our result. In the middle column, the putamen and caudate nucleus, which are important in the diagnosis of Parkinson's and Alzheimer's diseases \cite{broussolle1999relation,koivunen2008pet}, are most clearly depicted by our method. In the right column, the OSEM and R\textsuperscript{2}-Gaussian reconstructions show artifacts in the lower edge, while our reconstruction does not. Overall, among the three methods our method preserves image details best while exhibiting greater robustness to data noise in realistic PET data reconstruction.

\begin{table}
  \caption{Performance comparison on synthetic CT dataset.}
  \label{tab:synthetic}
  \small
  \setlength{\tabcolsep}{3pt}
  \centering
  \begin{tabular}{@{}l *{6}{c}@{}}  
    \toprule
    \multicolumn{1}{c}{Method} & \multicolumn{2}{c}{75 views} & \multicolumn{2}{c}{50 views} & \multicolumn{2}{c}{25 views} \\
    \cmidrule(r){1-1} \cmidrule(lr){2-3} \cmidrule(lr){4-5} \cmidrule(lr){6-7}
    & PSNR & SSIM & PSNR & SSIM & PSNR & SSIM \\
    \midrule
    R\textsuperscript{2}-Gaussian & 38.30 & 0.937 & 35.02 & 0.902 & 32.95 & \textbf{0.866}  \\
    Our method                    & \textbf{39.36} & 0.937 & \textbf{35.13} & 0.902 & \textbf{32.99} & 0.865 \\
    \bottomrule
  \end{tabular}
\end{table}

\begin{table}
  \caption{Performance comparison on real-world CT dataset.}
  \label{tab:real-world}
  \small
  \setlength{\tabcolsep}{3pt}
  \centering
  \begin{tabular}{@{}l *{6}{c}@{}}  
    \toprule
    \multicolumn{1}{c}{Method} & \multicolumn{2}{c}{75 views} & \multicolumn{2}{c}{50 views} & \multicolumn{2}{c}{25 views} \\
    \cmidrule(r){1-1} \cmidrule(lr){2-3} \cmidrule(lr){4-5} \cmidrule(lr){6-7}
    & PSNR & SSIM & PSNR & SSIM & PSNR & SSIM \\
    \midrule
    R\textsuperscript{2}-Gaussian & 35.71 & 0.928 & 34.22 & 0.915 & 30.56 & 0.852  \\
    Our method                    & \textbf{35.79} & 0.928 & \textbf{34.30} & 0.915 & \textbf{30.68} & \textbf{0.856} \\
    \bottomrule
  \end{tabular}
\end{table}

The CT reconstruction results on the synthetic and real-world datasets are summarized in \cref{tab:synthetic} and \cref{tab:real-world}, respectively. There is little difference in SSIM between R\textsuperscript{2}-Gaussian and our method. In terms of PSNR, our method achieves higher values than R\textsuperscript{2}-Gaussian in all tested cases. A paired t-test indicates that the PSNR improvement is statistically significant (p = 0.0048).

In practice, clinical CT is often interpreted qualitatively, and quantitative accuracy is typically less critical than in PET. Considering the trade-off between reconstruction quality and runtime, R\textsuperscript{2}-Gaussian may still be preferred for CT applications where speed is the dominant concern.

For PET reconstruction, our method demonstrates improved quantitative accuracy and greater flexibility to accommodate various scanner geometries, making it more suitable for quantitative PET tasks that require precise geometric modeling.

\section{Limitations and Future Works}
\label{sec:Limitations and Future Works}

\noindent\textbf{Limitations.} Unlike CT, there are few publicly available PET datasets, and PET data cannot be synthesized as straightforwardly as CT data because PET detects pairs of coincident gamma photons emitted in nearly 180° opposite directions within the imaged subject. Hence, in this study we relied on analytical and Monte Carlo simulations to approximate realistic measurements. High-fidelity Monte Carlo is extremely computationally expensive and therefore limited in scale, while analytical phantom simulations are cheap but do not capture many real-world effects, e.g., scatter, randoms. In addition, although the analytic integrals of 3D Gaussian ray tracing improve physical fidelity, a ray-tracing implementation is still computationally slower than highly parallel splatting-based implementations. This increases reconstruction time and limits throughput for large-scale experiments.

\noindent\textbf{Future works.} As a next step, we will extend the projection operator to explicitly model attenuation correction, scatter, and randoms within the ray-tracing pipeline, and to support Time-of-Flight (TOF) PET \cite{vandenberghe2016recent}. By incorporating these effects in a differentiable way, we expect this to move the method toward fully quantitative PET reconstruction \cite{lammertsma2017forward}. We will also investigate practical acceleration strategies, e.g., better BVH construction and mixed-precision computation, to reduce run time \cite{bavoil2007multi}.

Moreover, recent works show that integration with diffusion models or learned priors can further improve reconstruction quality \cite{mu2024gsd,liu20243dgs,zhou2024diffgs}. It would be interesting to explore combining diffusion-based regularization or generative priors with 3D Gaussian ray tracing to improve noise suppression and perceptual quality while keeping quantitative fidelity.

\section{Conclusion}

In this study, we propose a 3D Gaussian ray tracing framework for tomographic reconstruction. Our method avoids the local affine approximation used in splatting-based approaches by evaluating an analytic 3D Gaussian line integral along each ray. It is also adaptable to different scanner geometries, enabling PET reconstruction without resampling the sinogram and thus without compromising the original count data. Owing to the direct analytic integral and more accurate geometric modeling, our method demonstrates improved quantitative accuracy in PET imaging.
{
    \small
    \bibliographystyle{ieeenat_fullname}
    \bibliography{main}
}


\end{document}